\documentclass[10pt, a4paper]{article}

\usepackage{lrec-coling2024} 

\usepackage{subcaption}
\usepackage{graphicx}
\usepackage{booktabs}
\usepackage{multirow}
\usepackage{amsmath}
\newcommand{\wei}[2][]{\textcolor{black}{#2}}

\title{\vspace*{.5\baselineskip} \normalfont{ 
 \vspace*{.5\baselineskip} 
\textbf{Explaining Pre-Trained Language Models with Attribution Scores:\\
An Analysis in Low-Resource Settings}
}}

\name{Wei Zhou$^1$, Heike Adel$^3$, Hendrik Schuff$^4$, Ngoc Thang Vu$^2$} 

\address{$^1$Bosch Center for Artificial Intelligence, Renningen, Germany \\
         $^2$Institut f\"{u}r Maschinelle Sprachverarbeitung, University of Stuttgart, Germany \\
         $^3$Hochschule der Medien, Stuttgart, Germany\\
         $^4$Ubiquitous Knowledge Processing Lab, 
         Technical University of Darmstadt, Germany\\
         wei.zhou3@de.bosch.com, heike.adel@gmail.com, \\hendrik.schuff@tu-darmstadt.de, 
         thang.vu@ims.uni-stuttgart.de\\}

\abstract{
Attribution scores indicate the importance of different input parts and can, thus, explain model behaviour.
Currently, prompt-based models
are gaining popularity, i.a., due to their easier adaptability in low-resource settings. However, the quality of attribution scores extracted from prompt-based models has not been investigated yet.
In this work, we address this topic by analyzing attribution scores extracted from prompt-based models w.r.t.\ plausibility and faithfulness and comparing them with attribution scores extracted from fine-tuned models and large language models. 
In contrast to previous work, we introduce training size as another dimension into the analysis. 
We find that using the prompting paradigm (with either encoder-based or decoder-based models) yields more plausible explanations than fine-tuning the models in low-resource settings and Shapley Value Sampling consistently outperforms attention and Integrated Gradients in terms of plausibility and faithfulness scores.
 \\ \newline } 

\begin{document}

\maketitleabstract

\section{Introduction}
Recently, two paradigms of using 
pre-trained transformer models, such as BERT or GPT-2 \citep{Devlin2019BERTPO, Brown2020LanguageMA},
have gained popularity: \emph{fine-tuning} which adapts the weights of the model using task-specific training data, and \emph{prompting} which defines or learns so-called prompts to retrieve knowledge from the model, often leaving the model's weights unchanged. 

When deploying pre-trained models in real-world downstream applications, two challenges arise: (i) the need for \emph{explaining the results} as the models are very complex \citep{Madsen2021PosthocIF}, and (ii) the need for adapting the models in \emph{low-resource scenarios} as applications in special domains or languages typically do not provide many labeled training instances \citep{hedderich-etal-2021-survey}. 



For challenge (ii), previous work has shown that fine-tuning models in low-resource settings is hard (or even impossible for zero-resource settings) while prompting can yield good performance in those cases \citep{Brown2020LanguageMA,Schick2021ExploitingCF,Liu2021PretrainPA}.
In terms of challenge (i), there is a research gap of carefully analyzing the difference of fine-tuned models (FTMs) and prompt-based models (PBMs).
Most methods that have been proposed to enhance models' explainability \citep{Ribeiro2016WhySI, Lundberg2017AUA} have so far only been studied in the context of FTMs \citep{Atanasova2020ADS,DeYoung2020ERASERAB,Ding2021EvaluatingSM}, e.g., to answer the question which attribution method works best for different models and tasks.  
To the best of our knowledge, no previous work has explored attribution scores from PBMs (neither encoder-based nor decoder-based models, i.a., large language models) nor compared their quality to signals extracted from FTMs.

In this paper, we thus address the following questions: (1) \emph{How plausible and faithful are explanatory signals extracted from PBMs in comparison to FTMs?}  \wei[]{While plausibility shows how plausible an explanation is according to human understanding, faithfulness measures to what extent the deemed important tokens are truly important for the predictions of the model.
Thus, we evaluate explanations both from the perspective of humans and models, making the analysis comprehensive.\footnote{Those two dimensions are also commonly studied in related work on models' explainability \citep{Atanasova2020ADS,Ding2021EvaluatingSM}.} 
}
In addition, we introduce a \emph{new dimension into the analysis}, namely the number of training samples 
in order to carefully investigate the behaviour of different methods in low-resource settings. 

\wei{In our second research question, we investigate the effects of different attribution methods:} (2) \emph{How well do different attribution methods perform in terms of plausibility and faithfulness?} We answer this question by comparing different methods (namely attention, Integrated Gradients and Shapley Value Sampling) using extensive statistical significance tests. We focus on explanations in the form of attribution scores that highlight the importance of different input parts since they are more closely related to the model input and output than, e.g., generated free-text explanations. 

\wei{Our third question concerns the choice of the underlying model, taking into account the new trend of using large language models:}
(3) \emph{Do the results for PBMs also hold for decoder-based large language models?} We show that we get comparable results when extracting attribution scores from a large language model. 

For the first time, our paper shows that prompt-based models yield more plausible explanations than fine-tuned models in low-resource settings and Shapley Value Sampling consistently outperforms attention and Integrated Gradients in terms of 
both plausibility and faithfulness scores.
Thus, prompting pre-trained (either encoder-based or decoder-based) transformer models is better in low-resource settings than fine-tuning them, not only in terms of task performance but also when extracting attribution scores as explanations. 

\section{Extraction of Attribution Scores}
We analyze attribution scores from three different kinds of models: encoder-only models (e.g., BERT and similar models) following either the prompt-based paradigm (called ``PBMs'' in the following) or the fine-tuning paradigm (called ``FTMs''), and decoder-only models (e.g., large language models) following the prompt-based paradigm (called ``LLMs''). \wei[]{We do not investigate encoder-decoder models as we want to avoid mixing effects from cross-attention and self-attention.} In the following paragraphs, we describe how we extract attribution scores from the different model types.

\begin{figure}[t]
    \centering
    \includegraphics[width=1\columnwidth]{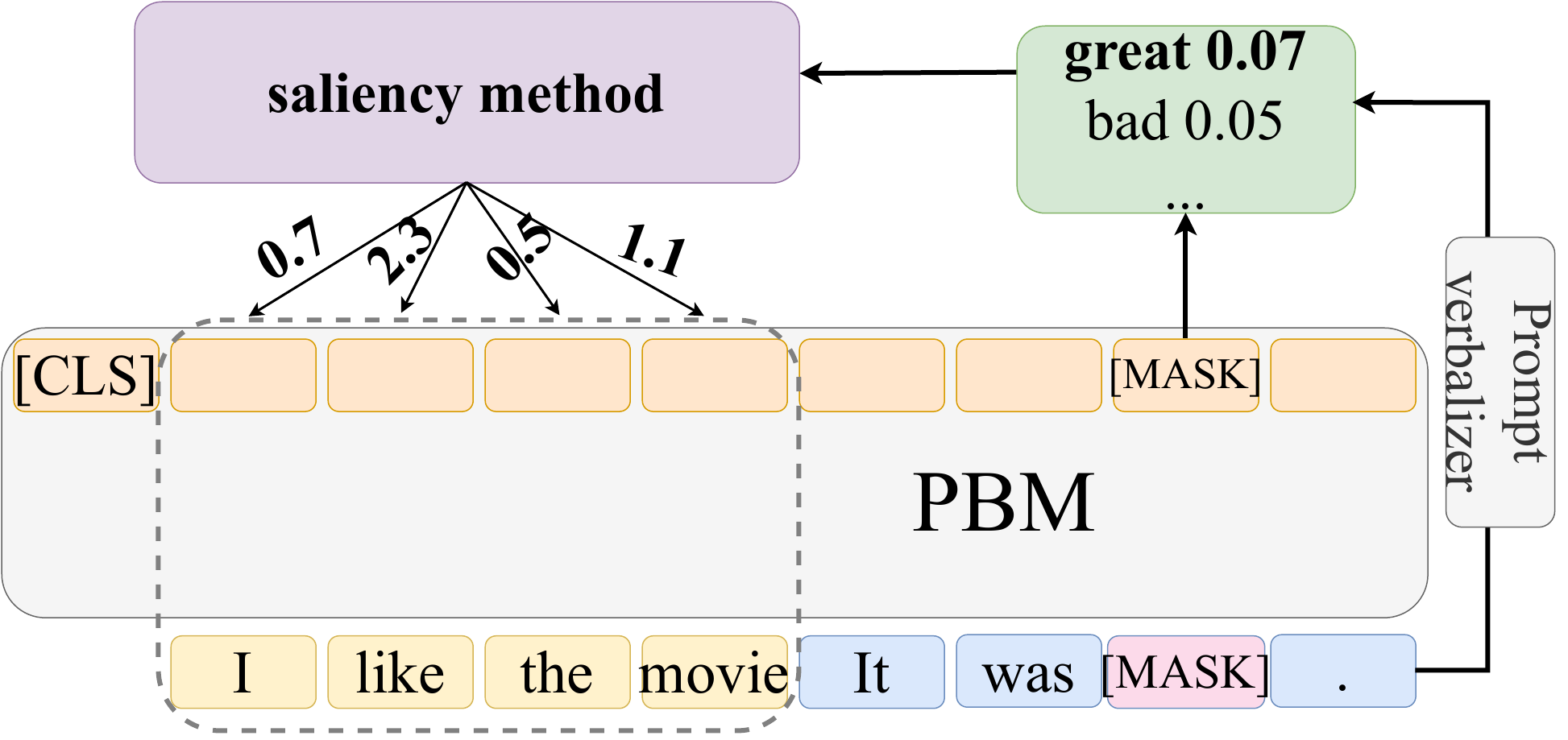}

    \caption{Extraction of explanatory signals from PBMs. Yellow boxes: \wei[input sentence]{actual task input}. Blue boxes: \wei[prompt]{trigger tokens. Pink box: prediction token}. Orange boxes: last hidden representations of PBM. Green box: predicted label (converted by verbalizer, e.g., positive $\rightarrow$ great, negative $\rightarrow$ bad).}
    \label{fig:PBM}
\end{figure}

\paragraph{Extraction from PBMs.}
We illustrate our approach for extracting attribution scores from encoder-only PBMs in Figure \ref{fig:PBM}: The input \wei[prompt]{} is composed of \wei[]{the actual task input (yellow boxes)}, trigger tokens (i.e., tokens providing task information\wei[]{, shown in blue boxes}), and a prediction token (i.e., the token that the model needs to predict to solve the task\wei[]{, shown in the pink box}). \wei[and the actual task input]{}
Based on the input\wei[prompt]{}, the model computes probabilities for the \wei[possible output candidates]{prediction token}. 
Given the predicted label tokens,\footnote{We decided to use the tokens from the verbalizer instead of the true task labels as in \citet{Atanasova2020ADS} because it cannot be assumed to have access to the true labels in real-world scenarios.} we then extract attribution scores for the \wei[part of the input that corresponds to the input sentences]{actual task input}. In particular, we use attention scores, Integrated Gradients and Shapley Value Sampling in our study. For attention, we extract attention scores from the last hidden layer of the \textsc{[MASK]} token, average them across different attention heads and normalize the attention scores over the \wei[input sentence]{actual task input}. \wei[]{For Integrated Gradients and Shapley Value Sampling, we calculate attribution scores using the Captum package.\footnote{https://captum.ai}}

\paragraph{Extraction from FTMs.}
For FTMs, the process is similar except that there are no prompts appended at the end of the sentences. Instead of using the language modeling head \wei[]{(the \textsc{[MASK]} token)} for prediction, we use the default classification head \wei[]{(the \textsc{[CLS]} token)} for FTMs and extract attribution scores for each \wei[input token]{token of the actual task input} based on the predictions.

\paragraph{Extraction from LLMs.}
Extracting attribution scores from generative models is more challenging as they typically generate a whole sequence of output tokens and the position of the \wei[verbalized class label]{prediction token} is not clear. \wei[Therefore,]{To tackle this issue,} we explicitly prompt the model to output only the verbalized class label.\footnote{We chose the verbalizer such that the label name is part of the model's vocabulary and is not split into several subtokens.} \wei[]{Prompts can be found in Section \ref{appendix:llm_prompts}.} \wei[Thus,]{Then} we \wei[can]{detect if the generated output corresponds to one of the verbalized class labels or not. If yes, we treat the class label as the prediction token. If not, we treat the first token in the generated output sequence as the prediction token. Finally, we} extract attribution scores for \wei[each input token]{the actual task input} based on the \wei[last token from the input sequence, as the hidden representation of that token decides the first prediction]{
prediction token, as we did in the extraction from PBMs or FTMs}.

\section{Experimental Setup}

\paragraph{Tasks and data sets.}
We use a sentiment classification (Tweet Sentiment Extraction (TSE)\footnote{https://www.kaggle.com/c/tweet-sentiment-extraction}) and a natural language inference dataset (e-SNLI \citep{Camburu2018eSNLINL}) to cover tasks of different semantic depth and use their annotations of token-level explanations. \wei[]{Statistics of the datasets can be found in Table \ref{tab:datasets}.\footnote{For TSE, we exclude data with the neutral label because their annotated explanations are mostly the whole sentence.}} 
To create low-resource settings, we subsample the training sets into six low-resource sets, ranging from eight instances to the whole set.

\begin{table}[!h]
\centering

    \begin{tabular}{c|c|c}
    \toprule
      \textbf{Data set} & \textbf{TSE} & \textbf{e-SNLI}\\

      \midrule
      Train & 11931 & 549367\\
      Dev & 2983 & 9842\\
      Test & 1449 & 9825\\
        \bottomrule
    \end{tabular}
        \caption{Number of training, development, and test instances in TSE and e-SNLI.}
      
\label{tab:datasets}
\end{table}


\paragraph{Base models.} For our main analysis, we focus on state-of-the-art encoder-based transformer models since running large language models (LLMs) on all our evaluation setups would have been infeasible due to extensive computational costs. In particular, we use BERT-base \citep{Devlin2019BERTPO}, BERT-large, and RoBERTa-large \citep{Liu2019RoBERTaAR}. 
Nevertheless, we also perform a small comparative study with LLMs afterwards, namely with the Vicuna model \citep{vicuna2023}, a fine-tuned LLaMA version \citep{llama2023}. 

\paragraph{Prompting methods.}
In our study, we focus on discrete prompts because  they are more explainable than continuous prompts and also the standard input for LLMs.
 To be able to factor out possible differences stemming from the choice of the prompting method, we study three different methods: 
\emph{Manual}\wei[(]{} uses a prompt from \citet{Schick2021ExploitingCF} \wei[]{and fine-tunes all parameters of the model.} \emph{BitFit} \wei[(]{} uses the manual prompt but updates only the bias terms of the model during \wei[training]{fine-tuning \citep{logan-iv-etal-2022-cutting}}, and \emph{BFF}\wei[(]{} automatically searches for a prompt \citep{Gao2021MakingPL} \wei[]{and fine-tunes all parameters with that prompt}. 
\wei[]{Prompts and verbalizers are provided in Table \ref{tab: example_prompts}}.\footnote{For the LLM, we use manual prompts only.}

\begin{table*}[h!]
\centering

    \begin{tabular}{llll}
    \toprule
      \textbf{Task} & \centering\textbf{Prompt} & \textbf{Verbalizer}&\textbf{Setting}\\
 
      \midrule
      \multirow{2}*{TSE} & {[}S{]} It was {[}P{]}.            & terrible/great & Manual/Bitfit\\ 
      
      & This is {[}P{]}.  {[}S{]}             &ragged/soldiers & BFF\\
      \midrule 
      \multirow{2}*{e-SNLI} & {[}S1{]} ? \textbar  {[}P{]} ,  {[}S2{]}     &yes/no/maybe&Manual/Bitfit\\
      & {[}S1{]} . {[}P{]} , no , {[}S2{]} & alright/except/watch&BFF\\
      
      \bottomrule
    \end{tabular}

  
    \caption{Prompts for TSE and e-SNLI in different settings. [S] stands for the sentence ([S1] and [S2] are the premise and hypothesis respectively), and [P] is the prediction token. For TSE, the verbalizers correspond to \textit{positive/negative}. For e-SNLI, the verbalizers correspond to \textit{entailment/contradiction/neutral}.}
    \label{tab: example_prompts}
\end{table*}





\paragraph{Training details.}
 We use 4-fold cross-validation to tune both PBMs and FTMs. 
 The hyperparameters \wei[will be provided in the appendix of the final version of the paper]{can be found in Section \ref{appendix:hyper}}.

\paragraph{Evaluation metrics.}
We evaluate the plausibility and faithfulness of the explanatory signals. Those two dimensions allow to investigate explanations both from the perspective of humans and models. They are also commonly used in related work on explainability \citep{Atanasova2020ADS,Ding2021EvaluatingSM}.

\emph{Plausibility} indicates how plausible an explanation is according to human intuition. We quantify this with average precision \citep{Atanasova2020ADS}.\footnote{\wei[]{ sklearn.metrics.average\_precision\_score}}

\emph{Faithfulness} shows a model’s ability to accurately represent its reasoning process. In related work, an established way of quantifying this is measuring the performance decrease when masking the most salient words \citep{DeYoung2020ERASERAB,Atanasova2020ADS}. We follow \citet{Atanasova2020ADS} and create several dataset perturbations by masking
0, 10, 20, ..., 100\% of the tokens in the order of decreasing saliency. To calculate a single measure for faithfulness, the area under the threshold-performance curve (AUC) is used. However, this measure does not allow cross-model comparisons. Therefore, we normalize the AUC as the proportion of the area under the curve to the whole area \wei[]{(calculated as the highest possible performance multiplied by the number of thresholds)}. The lower the normalized score, the better the explanation is in faithfully showing the model's reasoning.

To investigate the \emph{statistical significance} of our results, we apply Kruskal-Wallis tests and Dunn's Tests for pairwise differences.

\section{Results}
\subsection{Comparing PBMs and FTMs}


 \begin{figure*}
     \centering
    \begin{subfigure}[b]{0.49\textwidth}
         \centering
         \includegraphics[width=\textwidth]{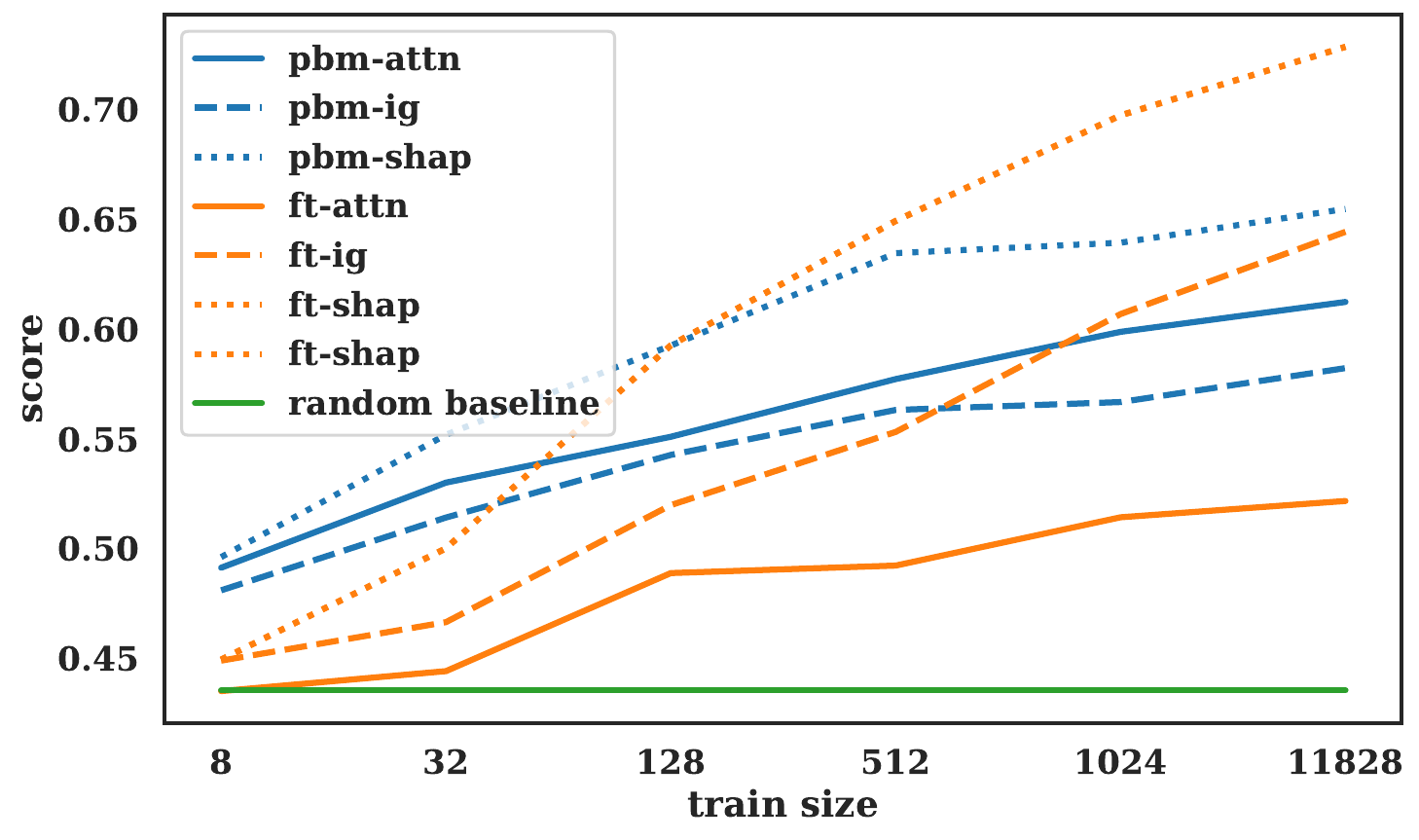}
    \caption{Plausibility Results on TSE.}
         \label{tse_pbm_ft}  
     \end{subfigure}
     \hfill
     \begin{subfigure}[b]{0.49\textwidth}
         \centering
         \includegraphics[width=\textwidth]{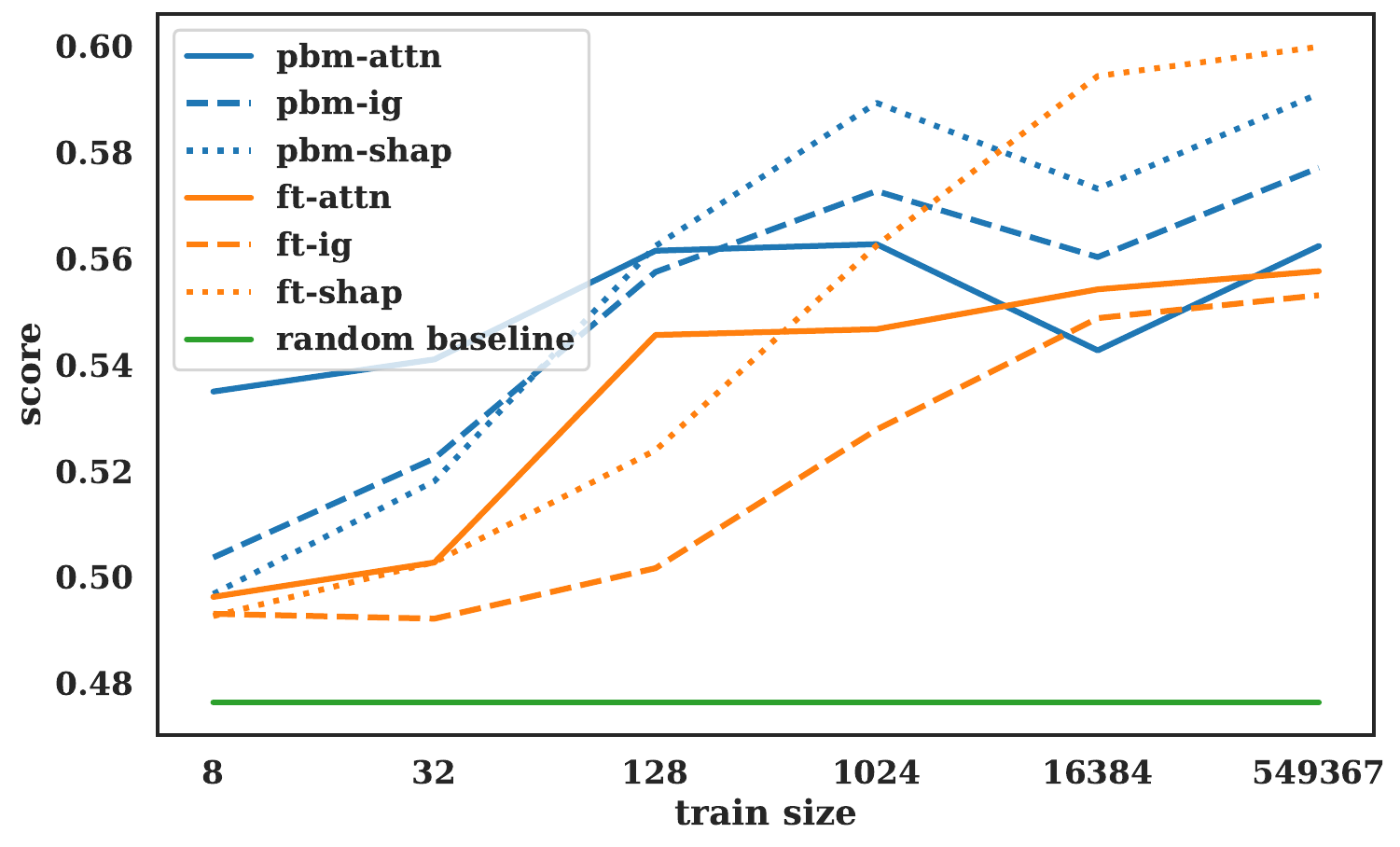}
         \caption{Plausibility Results on e-SNLI}
         \label{nli_pbm_ft}
     \end{subfigure}

   \begin{subfigure}[b]{0.49\textwidth}
         \centering
         \includegraphics[width=\textwidth]{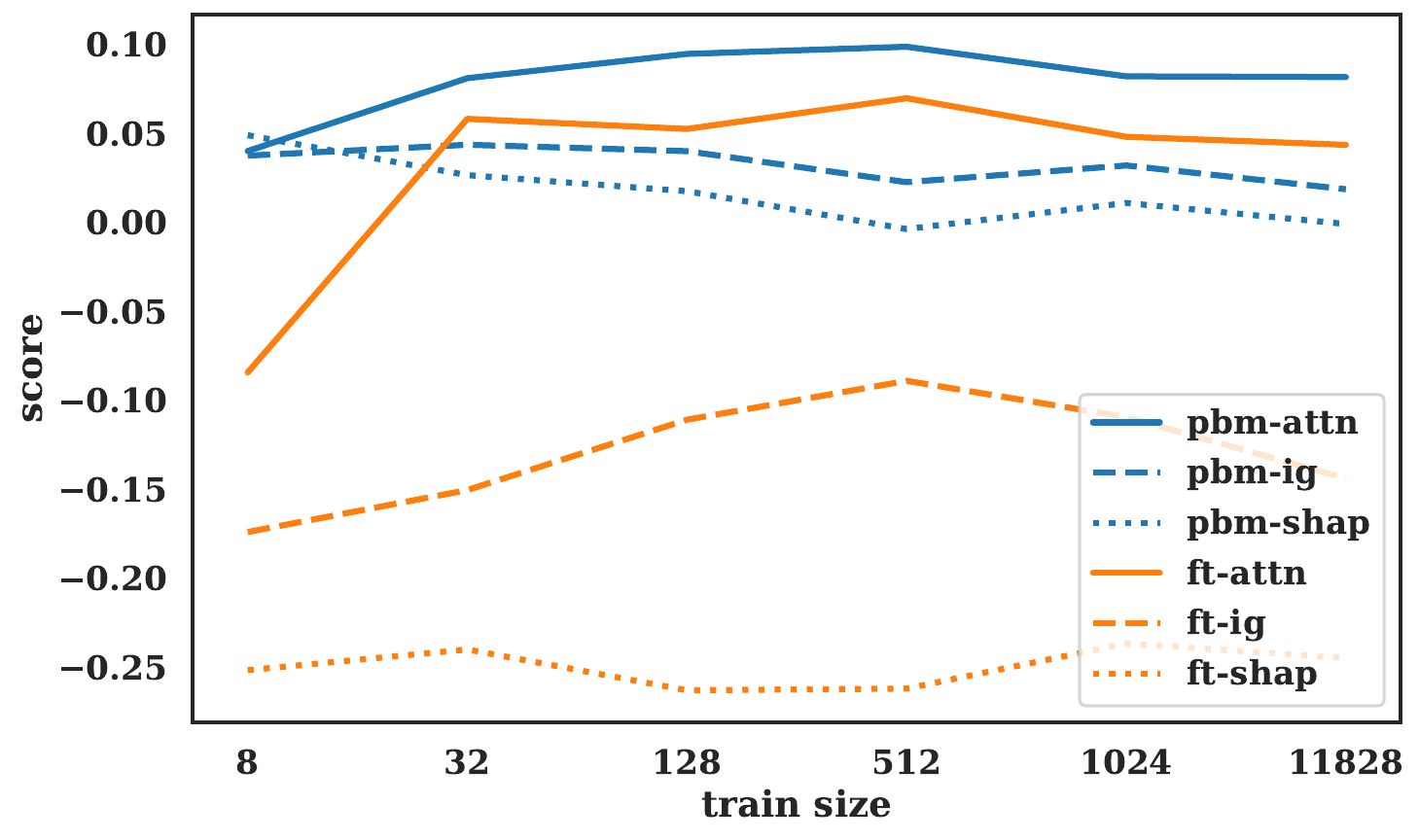}
         \caption{Faithfulness Results on TSE}
         \label{faith_tse_pbm_ft_}
     \end{subfigure}
     \hfill
     \begin{subfigure}[b]{0.49\textwidth}
         \centering
         \includegraphics[width=\textwidth]{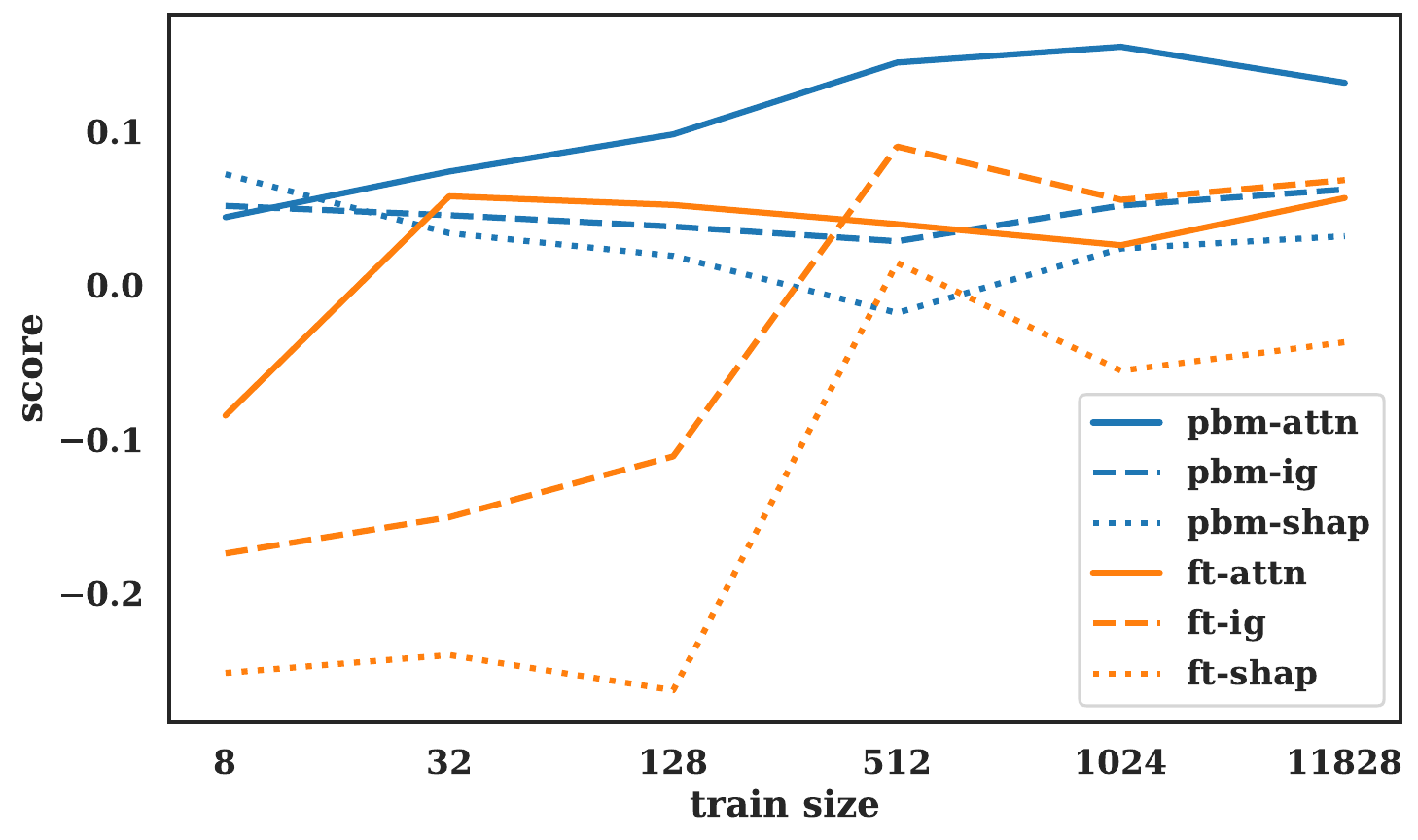}
         \caption{Faithfulness Results on e-SNLI}
         \label{faith_esni_pbm_ft}
     \end{subfigure}
     
        \caption{Plausibility (the higher the better) and faithfulness (the lower the better) scores for different prompting methods and fine-tuning and different explanation methods, averaged across base models and seeds. The faithfulness results are shown as the difference between faithfulness scores of the resp.\ explanation method and the gold standard. 
        \textit{attn}: attention, \textit{ig}: Integrated Gradients, \textit{shap}: ShapSample.}
        \label{pbm_ft}
\end{figure*}

We extract attribution scores from all PBMs and FTMs and compute its plausibility and faithfulness scores for different training sizes, averaging out effects from different prompting methods and base models. Figure \ref{pbm_ft} shows the results. Individual results for base models and attribution methods \wei[will be given in the appendix of the final version of the paper]{can be found in Section \ref{appendix:break_down}. We also report the task performance of the models in Section \ref{appendix:task_p}}.

\paragraph{Plausibility.}
For smaller training sizes, PBMs outperform FTMs but the trend reverses as the training size increases. To investigate whether the differences between PBMs and FTMs in the low/rich-resource settings are significant, we set up two bins for each task: we treat the two training sets with fewest data as low-resource and the two training sets with most data as high-resource.
Within the low-resource part of the data, we find all comparisons to be statistically significant (TSE: $H$(89)=73.86, $p$<0.001, e-SNLI: $H$(89)=29.24,  $p$<0.001). 
Within the high-resource part of the data, the differences are not significant.
We also calculate the random baseline for plausibility scores (0.436 for TSE and 0.476 for e-SNLI) and find that explanations provided by PBMs achieve considerably higher plausibility scores than the random baselines in low-resource setting.
Method-wise, we find that for both tasks, the plausibility scores of the explanations extracted by ShapSample are significantly higher than those from attention and Integrated Gradients. 

\paragraph{Plausibility error analysis.}
We sample 20 instances per dataset for each attribution method to conduct a small error analysis in terms of plausibility of explanations. We find that Integrated Gradients tend to assign negative values to functional words. We also find that attention seems to encode sentence information into a single token, so a specific token can get high attribution scores.

\paragraph{Faithfulness.}
Figures \ref{faith_tse_pbm_ft_} and \ref{faith_esni_pbm_ft} show that faithfulness scores are influenced by the attribution methods. For instance, explanations extracted from FTMs with ShapSample are more faithful than explanations from PBMs independent of the number of resources. Explanations from PBMs with attention lead to the lowest faithfulness scores across all training sizes. For both datasets, we observe significant differences for all attribution method pairs except for ShapSample and gold. 
Thus, Shapley Value Sampling attribution scores are comparably faithful as gold annotations.

\subsection{Studying LLMs}
Given the increased relevance of large language models, we now investigate whether our findings hold for them as well and which plausibility and faithfulness scores we get for them compared to PBMs (i.e., encoder-based models used with prompting). 
Due to the large computational costs for obtaining attribution scores from LLMs, we limit the number of test instances to 100 for each data set and evaluate the 8-shot setting only. For the LLM, the 8 training samples are provided in each input prompt. \wei[\footnote{The prompts we use for vicuna will be provided in the appendix of the final version of the paper.}]{The prompts can be found in Section \ref{appendix:llm_prompts}.} For the PBM (we chose RoBERTa-Large with BitFit prompts which was the best performing individual model in our previous analysis), the 8 training samples are used to tune the \wei[prompt]{bias terms of the model}.




\begin{table}
\resizebox{\columnwidth}{!}{%
  \begin{tabular}{llrrrrrr}
    \toprule \textbf{Data} &
    \textbf{Model} &
      \multicolumn{3}{c}{\textbf{Plausibility}} &
      \multicolumn{3}{c}{\textbf{Faithfulness}}
      \\
     & & {attn} & {ig} & {shap} & {attn} & {ig} & {shap} \\
      \midrule
      \multirow{2}{*}{TSE} &
    RoBERTa & .56 & .57& .56 & .02 & \textbf{.00} & .01 \\
    & Vicuna & .47 & .57 & \textbf{.59} & .07 & .06 & .02 \\
    \midrule
     \multirow{2}{*}{e-SNLI} &   RoBERTa & .53 & .51& .50 & .22 & .09 & .11 \\
    & Vicuna & .43 & .51 & \textbf{.55} & .02 & .05 & \textbf{.00} \\
    \bottomrule
  \end{tabular}%
 }
 \caption{Plausibility (the higher the better) and faithfulness (the lower the better) scores of explanations obtained from Vicuna and RoBERTa.}
  \label{vicuna-results}
\end{table}

The results in Table \ref{vicuna-results} show that Shapley Value Sampling again leads to more plausible and faithful explanations for Vicuna.
When comparing Vicuna with RoBERTa, we note larger performance gaps among the attribution methods. 
We further note that the plausibility scores of attention are even lower for Vicuna than for RoBERTa.
A reason could be that LLMs encode a larger input context
and, thus,
information of tokens that are irrelevant to the prediction might also be encoded.

\subsection{Discussion}

\paragraph{Comparison of attribution methods.}
ShapSample consistently yields more plausible explanations than methods. We assume the reason for this lies in the calculation of Shapley Values: it takes in every permutation of features enabled to calculate a feature’s importance. For instance, if we have a feature set {“good”, “day”}, the attribution score of the feature “good” is calculated by every permutation that contains it, i.e., “good” and “good day”. Whereas for Integrated Gradients, this is not considered. We think taking each permutation to calculate feature importance is helpful in models like BERT, as context is of vital importance. Attention is the least plausible; this observation is in line with previous works, e.g., \citet{bibal-etal-2022-attention}. 

\paragraph{PBMs vs.\ FTMs vs.\ LLMs in low-resource settings.}
PBMs yield more plausible attribution scores than FTMs in low-resource settings. We think this might be because PBMs pick up task information quicker than FTMs 
in the low-resource settings, so the explanations given by PBMs are more plausible. Our study with LLMs shows that the trends of LLMs are comparable to the trends of PMBs, indicating the relevancy of our findings.

\section{Conclusion}
In this paper, we studied attribution scores extracted from prompt-based models in comparison to fine-tuned models, and compared different attribution methods w.r.t.\ plausibility and faithfulness scores. Our main findings were: (1) Prompt-based models generate more plausible explanations in low-resource settings. (2) Shapley Value Sampling outperforms other attribution methods, such as attention and Integrated Gradients across tasks and settings and is similarly faithful as gold annotations. (3) Our findings seem to be transferable to generative large language models.

Directions for future work are the investigation of soft prompts as well as a more extensive study of explanatory signals from large language models. 

\section{Bibliographical References}
\bibliographystyle{lrec-coling2024-natbib}
\bibliography{lrec-coling2024-example}

\begin{thebibliography}{0}
\expandafter\ifx\csname natexlab\endcsname\relax\def\natexlab#1{#1}\fi

\end{thebibliography}


\begin{thebibliography}{18}
\expandafter\ifx\csname natexlab\endcsname\relax\def\natexlab#1{#1}\fi

\bibitem[{Atanasova et~al.(2020)Atanasova, Simonsen, Lioma, and
  Augenstein}]{Atanasova2020ADS}
Pepa Atanasova, Jakob~Grue Simonsen, Christina Lioma, and Isabelle Augenstein.
  2020.
\newblock \href {https://doi.org/10.18653/v1/2020.emnlp-main.263} {A diagnostic
  study of explainability techniques for text classification}.
\newblock In \emph{Proceedings of the 2020 Conference on Empirical Methods in
  Natural Language Processing (EMNLP)}, pages 3256--3274, Online. Association
  for Computational Linguistics.

\bibitem[{Bibal et~al.(2022)Bibal, Cardon, Alfter, Wilkens, Wang,
  Fran{\c{c}}ois, and Watrin}]{bibal-etal-2022-attention}
Adrien Bibal, R{\'e}mi Cardon, David Alfter, Rodrigo Wilkens, Xiaoou Wang,
  Thomas Fran{\c{c}}ois, and Patrick Watrin. 2022.
\newblock \href {https://doi.org/10.18653/v1/2022.acl-long.269} {Is attention
  explanation? an introduction to the debate}.
\newblock In \emph{Proceedings of the 60th Annual Meeting of the Association
  for Computational Linguistics (Volume 1: Long Papers)}, pages 3889--3900,
  Dublin, Ireland. Association for Computational Linguistics.

\bibitem[{Brown et~al.(2020)Brown, Mann, Ryder, Subbiah, Kaplan, Dhariwal,
  Neelakantan, Shyam, Sastry, Askell, Agarwal, Herbert-Voss, Krueger, Henighan,
  Child, Ramesh, Ziegler, Wu, Winter, Hesse, Chen, Sigler, Litwin, Gray, Chess,
  Clark, Berner, McCandlish, Radford, Sutskever, and
  Amodei}]{Brown2020LanguageMA}
Tom~B. Brown, Benjamin Mann, Nick Ryder, Melanie Subbiah, Jared Kaplan,
  Prafulla Dhariwal, Arvind Neelakantan, Pranav Shyam, Girish Sastry, Amanda
  Askell, Sandhini Agarwal, Ariel Herbert-Voss, Gretchen Krueger, Tom Henighan,
  Rewon Child, Aditya Ramesh, Daniel~M. Ziegler, Jeffrey Wu, Clemens Winter,
  Christopher Hesse, Mark Chen, Eric Sigler, Mateusz Litwin, Scott Gray,
  Benjamin Chess, Jack Clark, Christopher Berner, Sam McCandlish, Alec Radford,
  Ilya Sutskever, and Dario Amodei. 2020.
\newblock Language models are few-shot learners.
\newblock In \emph{Proceedings of the 34th International Conference on Neural
  Information Processing Systems}, NIPS'20, Red Hook, NY, USA. Curran
  Associates Inc.

\bibitem[{Camburu et~al.(2018)Camburu, Rockt{\"a}schel, Lukasiewicz, and
  Blunsom}]{Camburu2018eSNLINL}
Oana-Maria Camburu, Tim Rockt{\"a}schel, Thomas Lukasiewicz, and Phil Blunsom.
  2018.
\newblock e-snli: Natural language inference with natural language
  explanations.
\newblock In \emph{NeurIPS}.

\bibitem[{Chiang et~al.(2023)Chiang, Li, Lin, Sheng, Wu, Zhang, Zheng, Zhuang,
  Zhuang, Gonzalez, Stoica, and Xing}]{vicuna2023}
Wei-Lin Chiang, Zhuohan Li, Zi~Lin, Ying Sheng, Zhanghao Wu, Hao Zhang, Lianmin
  Zheng, Siyuan Zhuang, Yonghao Zhuang, Joseph~E. Gonzalez, Ion Stoica, and
  Eric~P. Xing. 2023.
\newblock \href {https://lmsys.org/blog/2023-03-30-vicuna/} {Vicuna: An
  open-source chatbot impressing gpt-4 with 90\%* chatgpt quality}.

\bibitem[{Devlin et~al.(2019)Devlin, Chang, Lee, and
  Toutanova}]{Devlin2019BERTPO}
Jacob Devlin, Ming-Wei Chang, Kenton Lee, and Kristina Toutanova. 2019.
\newblock \href {https://doi.org/10.18653/v1/N19-1423} {{BERT}: Pre-training of
  deep bidirectional transformers for language understanding}.
\newblock In \emph{Proceedings of the 2019 Conference of the North {A}merican
  Chapter of the Association for Computational Linguistics: Human Language
  Technologies, Volume 1 (Long and Short Papers)}, pages 4171--4186,
  Minneapolis, Minnesota. Association for Computational Linguistics.

\bibitem[{DeYoung et~al.(2020)DeYoung, Jain, Rajani, Lehman, Xiong, Socher, and
  Wallace}]{DeYoung2020ERASERAB}
Jay DeYoung, Sarthak Jain, Nazneen~Fatema Rajani, Eric Lehman, Caiming Xiong,
  Richard Socher, and Byron~C. Wallace. 2020.
\newblock \href {https://doi.org/10.18653/v1/2020.acl-main.408} {{ERASER}: {A}
  benchmark to evaluate rationalized {NLP} models}.
\newblock In \emph{Proceedings of the 58th Annual Meeting of the Association
  for Computational Linguistics}, pages 4443--4458, Online. Association for
  Computational Linguistics.

\bibitem[{Ding and Koehn(2021)}]{Ding2021EvaluatingSM}
Shuoyang Ding and Philipp Koehn. 2021.
\newblock \href {https://doi.org/10.18653/v1/2021.naacl-main.399} {Evaluating
  saliency methods for neural language models}.
\newblock In \emph{Proceedings of the 2021 Conference of the North American
  Chapter of the Association for Computational Linguistics: Human Language
  Technologies}, pages 5034--5052, Online. Association for Computational
  Linguistics.

\bibitem[{Gao et~al.(2021)Gao, Fisch, and Chen}]{Gao2021MakingPL}
Tianyu Gao, Adam Fisch, and Danqi Chen. 2021.
\newblock \href {https://doi.org/10.18653/v1/2021.acl-long.295} {Making
  pre-trained language models better few-shot learners}.
\newblock In \emph{Proceedings of the 59th Annual Meeting of the Association
  for Computational Linguistics and the 11th International Joint Conference on
  Natural Language Processing (Volume 1: Long Papers)}, pages 3816--3830,
  Online. Association for Computational Linguistics.

\bibitem[{Hedderich et~al.(2021)Hedderich, Lange, Adel, Str{\"o}tgen, and
  Klakow}]{hedderich-etal-2021-survey}
Michael~A. Hedderich, Lukas Lange, Heike Adel, Jannik Str{\"o}tgen, and
  Dietrich Klakow. 2021.
\newblock \href {https://doi.org/10.18653/v1/2021.naacl-main.201} {A survey on
  recent approaches for natural language processing in low-resource scenarios}.
\newblock In \emph{Proceedings of the 2021 Conference of the North American
  Chapter of the Association for Computational Linguistics: Human Language
  Technologies}, pages 2545--2568, Online. Association for Computational
  Linguistics.

\bibitem[{Liu et~al.(2022)Liu, Yuan, Fu, Jiang, Hayashi, and
  Neubig}]{Liu2021PretrainPA}
Pengfei Liu, Weizhe Yuan, Jinlan Fu, Zhengbao Jiang, Hiroaki Hayashi, and
  Graham Neubig. 2022.
\newblock \href {https://doi.org/10.1145/3560815} {Pre-train, prompt, and
  predict: A systematic survey of prompting methods in natural language
  processing}.
\newblock \emph{ACM Comput. Surv.}

\bibitem[{Liu et~al.(2019)Liu, Ott, Goyal, Du, Joshi, Chen, Levy, Lewis,
  Zettlemoyer, and Stoyanov}]{Liu2019RoBERTaAR}
Yinhan Liu, Myle Ott, Naman Goyal, Jingfei Du, Mandar Joshi, Danqi Chen, Omer
  Levy, Mike Lewis, Luke Zettlemoyer, and Veselin Stoyanov. 2019.
\newblock Roberta: A robustly optimized bert pretraining approach.
\newblock \emph{ArXiv}, abs/1907.11692.

\bibitem[{Logan~IV et~al.(2022)Logan~IV, Balazevic, Wallace, Petroni, Singh,
  and Riedel}]{logan-iv-etal-2022-cutting}
Robert Logan~IV, Ivana Balazevic, Eric Wallace, Fabio Petroni, Sameer Singh,
  and Sebastian Riedel. 2022.
\newblock \href {https://doi.org/10.18653/v1/2022.findings-acl.222} {Cutting
  down on prompts and parameters: Simple few-shot learning with language
  models}.
\newblock In \emph{Findings of the Association for Computational Linguistics:
  ACL 2022}, pages 2824--2835, Dublin, Ireland. Association for Computational
  Linguistics.

\bibitem[{Lundberg and Lee(2017)}]{Lundberg2017AUA}
Scott~M. Lundberg and Su-In Lee. 2017.
\newblock A unified approach to interpreting model predictions.
\newblock In \emph{Proceedings of the 31st International Conference on Neural
  Information Processing Systems}, NIPS'17, page 4768–4777, Red Hook, NY,
  USA. Curran Associates Inc.

\bibitem[{Madsen et~al.(2022)Madsen, Reddy, and Chandar}]{Madsen2021PosthocIF}
Andreas Madsen, Siva Reddy, and Sarath Chandar. 2022.
\newblock \href {https://doi.org/10.1145/3546577} {Post-hoc interpretability
  for neural nlp: A survey}.
\newblock \emph{ACM Comput. Surv.}

\bibitem[{Ribeiro et~al.(2016)Ribeiro, Singh, and Guestrin}]{Ribeiro2016WhySI}
Marco~Tulio Ribeiro, Sameer Singh, and Carlos Guestrin. 2016.
\newblock "why should i trust you?": Explaining the predictions of any
  classifier.
\newblock \emph{Proceedings of the 22nd ACM SIGKDD International Conference on
  Knowledge Discovery and Data Mining}.

\bibitem[{Schick and Sch{\"u}tze(2021)}]{Schick2021ExploitingCF}
Timo Schick and Hinrich Sch{\"u}tze. 2021.
\newblock \href {https://doi.org/10.18653/v1/2021.eacl-main.20} {Exploiting
  cloze-questions for few-shot text classification and natural language
  inference}.
\newblock In \emph{Proceedings of the 16th Conference of the European Chapter
  of the Association for Computational Linguistics: Main Volume}, pages
  255--269, Online. Association for Computational Linguistics.

\bibitem[{Touvron et~al.(2023)Touvron, Lavril, Izacard, Martinet, Lachaux,
  Lacroix, Rozi\`{e}re, Goyal, Hambro, Azhar, Rodriguez, Joulin, Grave, and
  Lample}]{llama2023}
Hugo Touvron, Thibaut Lavril, Gautier Izacard, Xavier Martinet, Marie-Anne
  Lachaux, Timoth\'{e}e Lacroix, Baptiste Rozi\`{e}re, Naman Goyal, Eric
  Hambro, Faisal Azhar, Aurelien Rodriguez, Armand Joulin, Edouard Grave, and
  Guillaume Lample. 2023.
\newblock {LLaMA: Open} and efficient foundation language models.
\newblock \emph{ArXiv}, abs/2302.13971.

\end{thebibliography}

\bibliographystylelanguageresource{lrec-coling2024-natbib}
\bibliographylanguageresource{languageresource}

\appendix
\section{Appendix}
\label{sec:appendix}

 
      
  

\subsection{Hyperparameters}
\label{appendix:hyper}
Table \ref{tab:hyper} provides the hyperparameters for training our models. The learning rate and accumulation steps are automatically selected based on the validation accuracy from the range of [1, 2, 4] for the accumulation steps and [1.0e-5, 5.0e-5, 1.0e-4, 5.0e-4] for the learning rate. 

\begin{table}[h!]
\centering

    \begin{tabular}{c |c|c|c}
    \toprule
    \textbf{Task}&
      \textbf{Training size} & \textbf{Batch size} & \textbf{Epoch}\\
      \midrule
      TSE & 8/32/128 & 2 & 30\\
     TSE & 512 & 4 & 15\\
      TSE & 2048/11828 & 8 & 5\\
        \midrule
    e-SNLI & 8/32/128 & 2 & 30\\
      e-SNLI & 1024 & 4 & 15\\
      e-SNLI & 16384/549367 & 32& 3\\
      \bottomrule
    \end{tabular}
        \caption{Hyperparameters.}
\label{tab:hyper}
\end{table}

\subsection{LLM Prompts}
\label{appendix:llm_prompts}
\begin{itemize}
    \item TSE: You will be given a target sentence and you will decide the sentiment of the sentence (Please return either yes or no only). Here are some examples: Input: \{s1\} Output: \{l1\}...Input: \{s8\} Output: \{l8\}
    \item e-SNLI: You will be given a pair of sentences and you will decide the relationship between the sentences (Please return yes for entailment, no for contradiction, maybe for neutral only.). Here are some examples:  Input1: \{s1\} Input2: \{s2\} Output: \{l1\}...Input1: \{s8\} Input2: \{s8\} Output: \{l8\}
\end{itemize}

\begin{figure}
\centering
\includegraphics[width=\columnwidth]{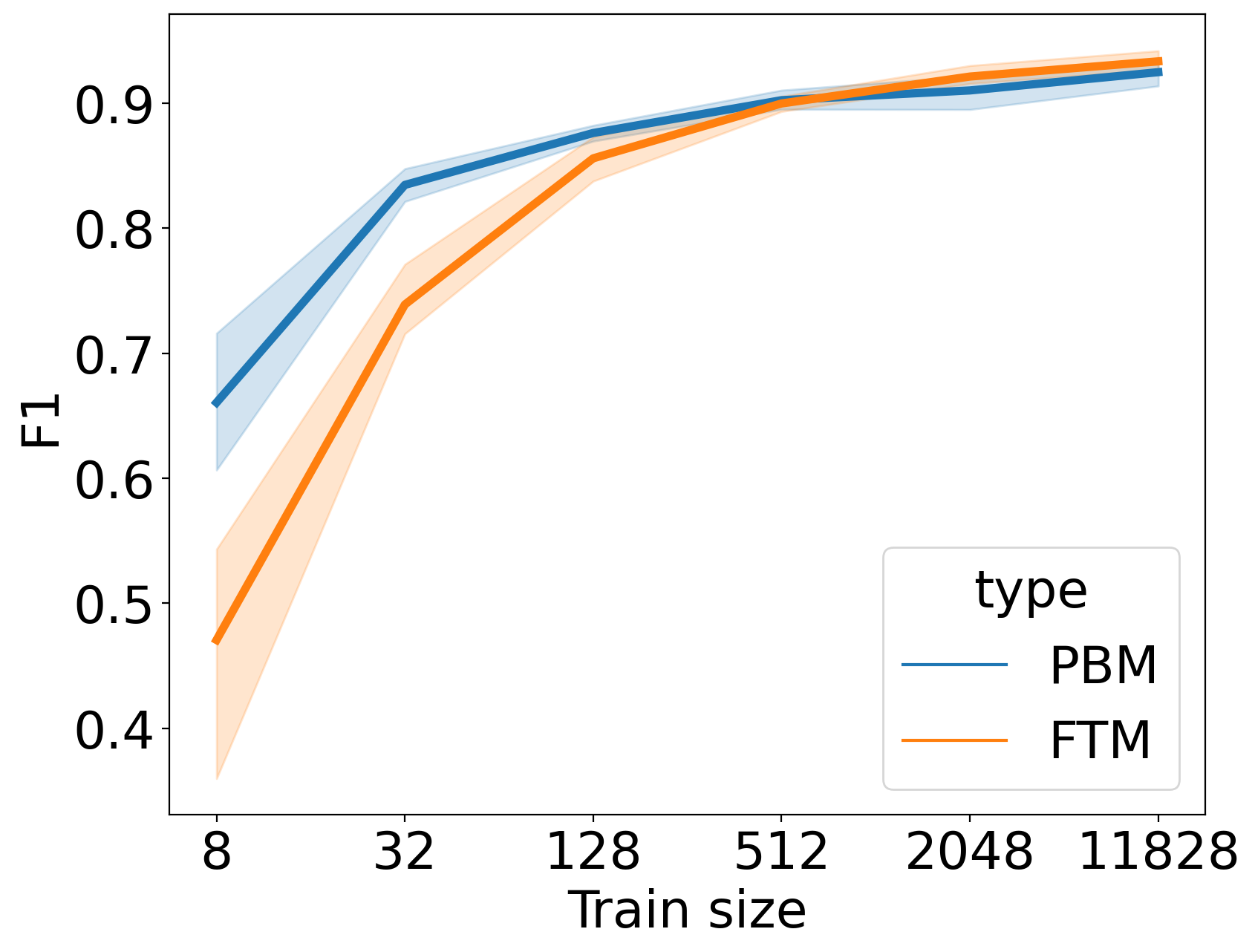}

\medskip
\includegraphics[width=\columnwidth]{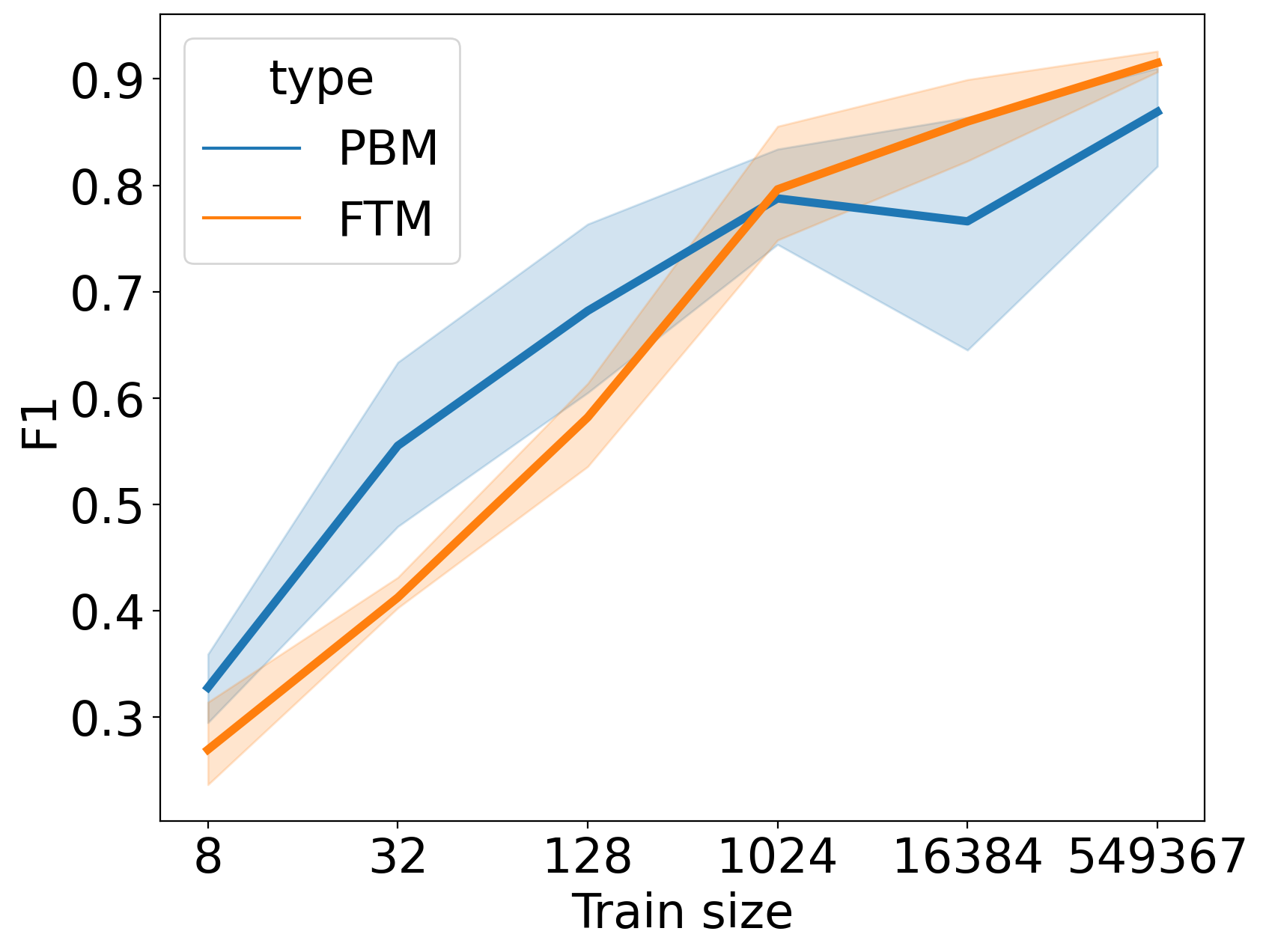}

 \caption{The $F_{1}$ scores of models trained with different sizes. From top to bottom: TSE and e-SNLI.}
 \label{tab:task_p}
\end{figure}

\subsection{Task Performance}
\label{appendix:task_p}
Figure \ref{tab:task_p} shows the task performances of PBMs and FTMs with regards to different training sizes.

\subsection{Comparing Saliency Methods and Base Models}
\label{appendix:break_down}

\begin{figure*}
     \centering
     \begin{subfigure}[b]{0.48\textwidth}
         \centering
         \includegraphics[width=\textwidth]{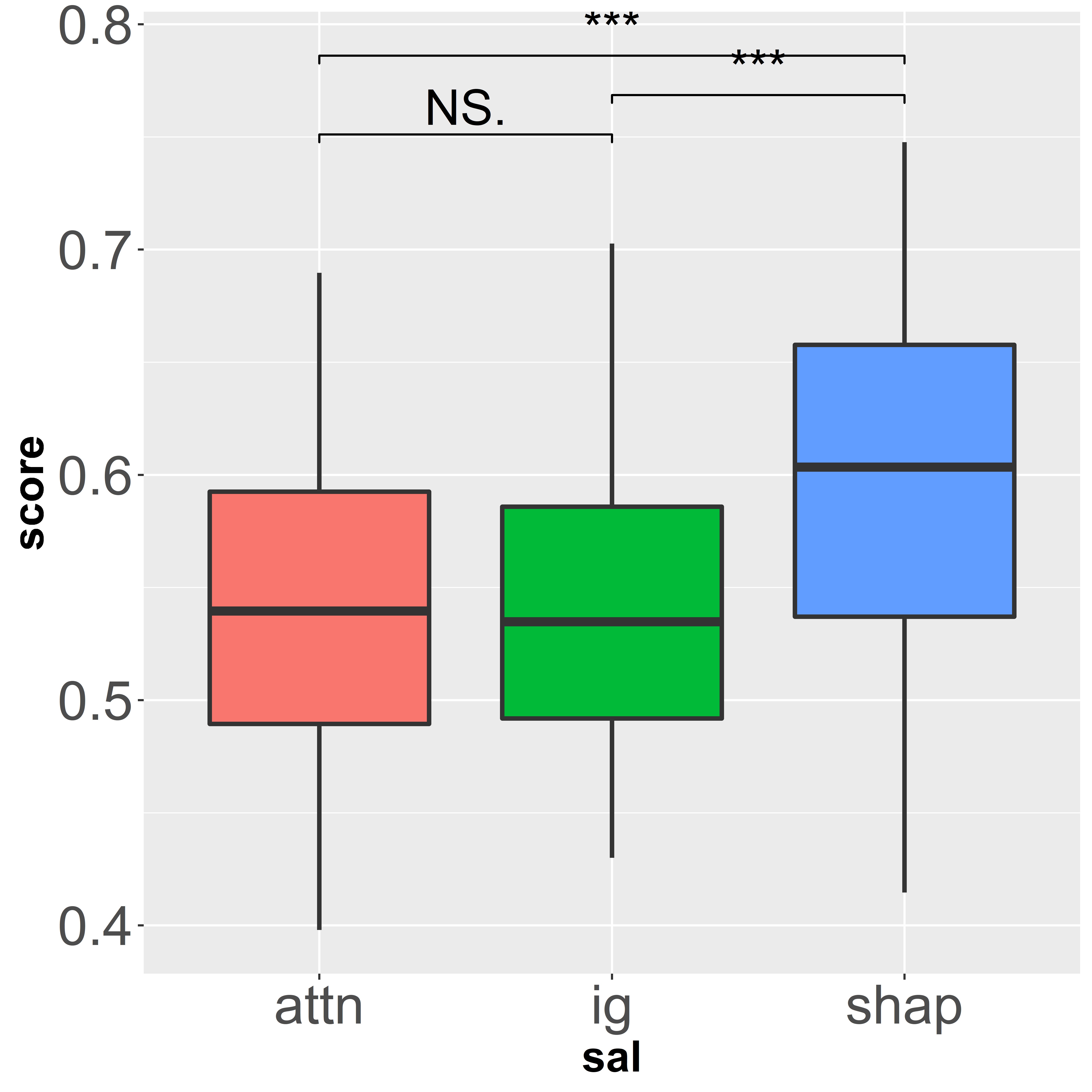}
         \caption{Plausibility:TSE}
         \label{tse_size}
     \end{subfigure}
     \hfill
     \begin{subfigure}[b]{0.48\textwidth}
         \centering
         \includegraphics[width=\textwidth]{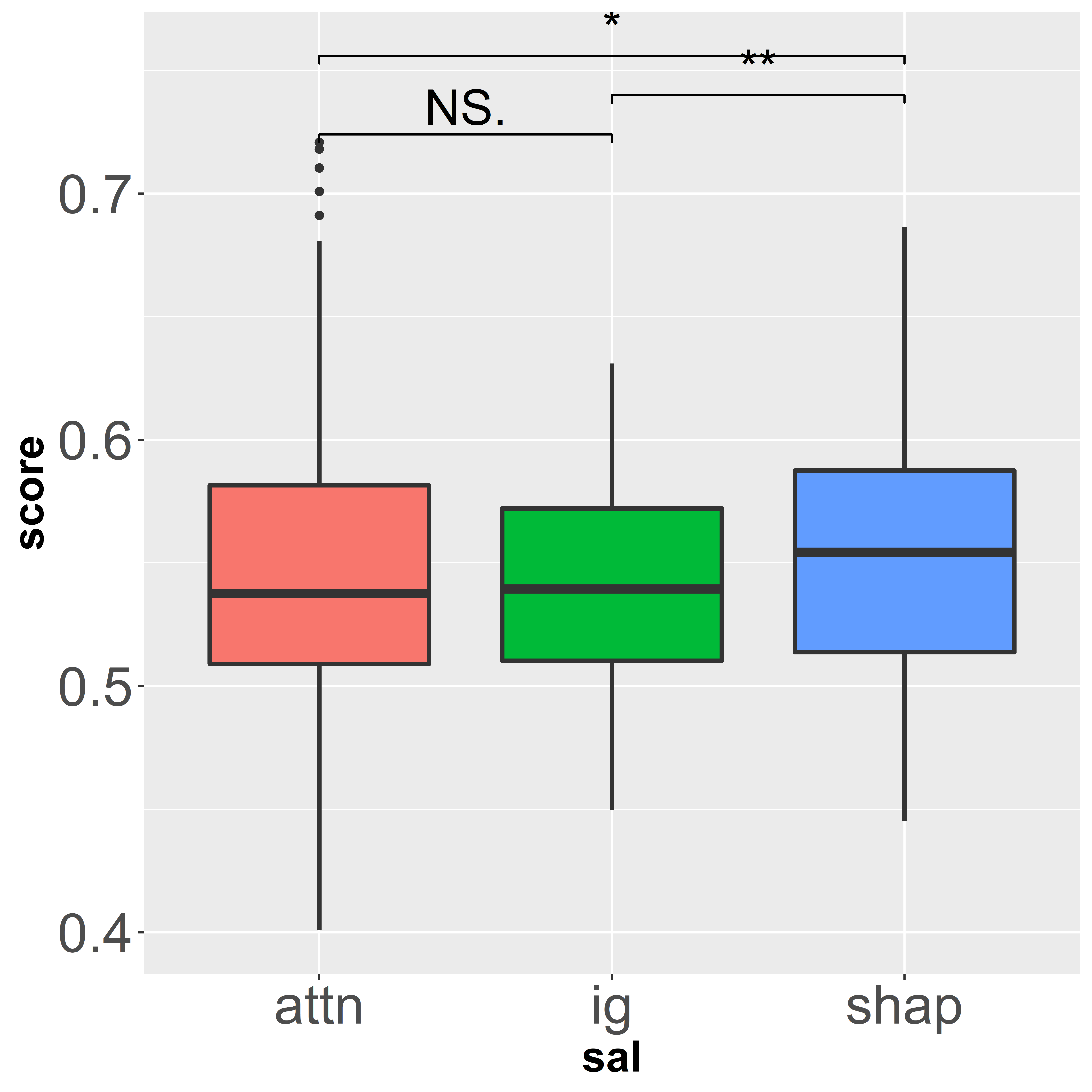}
         \caption{Plausibility:e-SNLI}
         \label{}
     \end{subfigure}

   \begin{subfigure}[b]{0.48\textwidth}
         \centering
         \includegraphics[width=\textwidth]{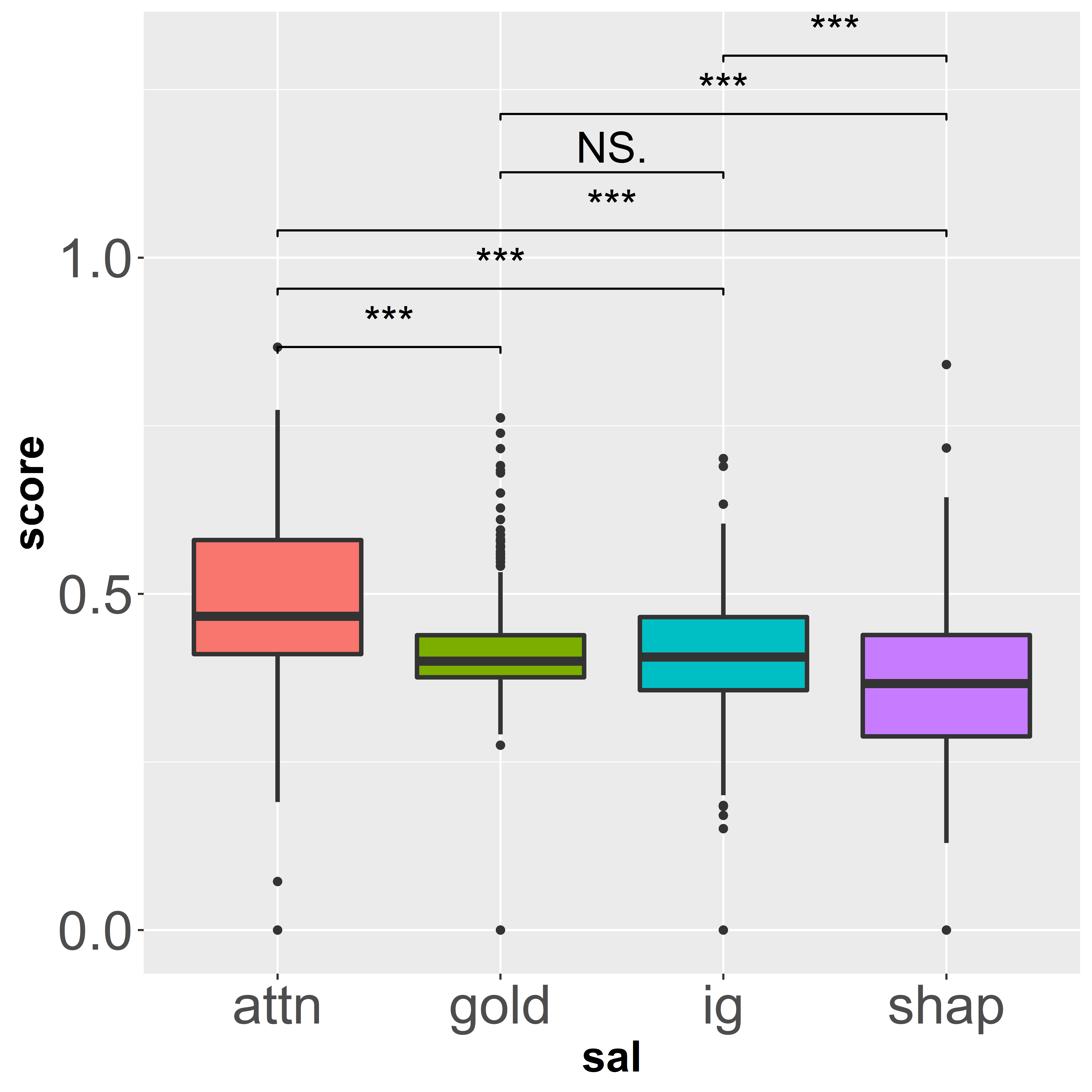}
         \caption{Faithfulness:TSE}
         \label{nli_size}
     \end{subfigure}
     \hfill
     \begin{subfigure}[b]{0.48\textwidth}
         \centering
         \includegraphics[width=\textwidth]{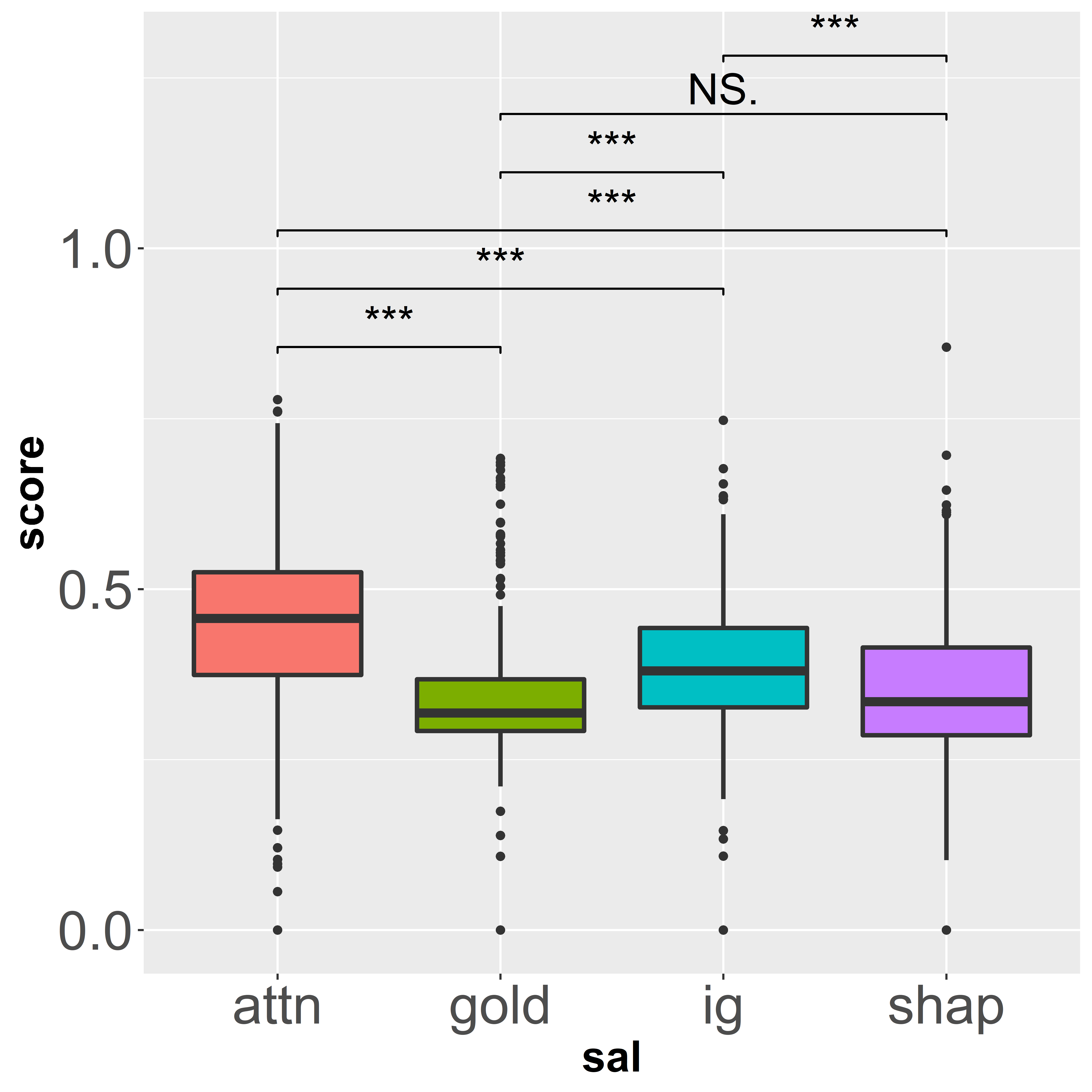}
         \caption{Faithfulness:e-SNLI}
         \label{nli_prompt}
     \end{subfigure}
     
  \caption{The Plausibility scores of explanatory signals, averaged across base models, training sizes and prompting methods. \textit{attn} stands for attention. \textit{ig} stands for Integrated Gradients and \textit{shap} stands for Shapley Value Sampling. \textit{gold} stands for the gold annotations. \textit{NS} stands for the no significant difference. \textit{*, **, ***} stand for \textit{p}-value <.05, .01 and .001.}
    \label{compare_sal}
\end{figure*}

\begin{figure*}
     \centering
     \begin{subfigure}[b]{0.48\textwidth}
         \centering
         \includegraphics[width=\textwidth]{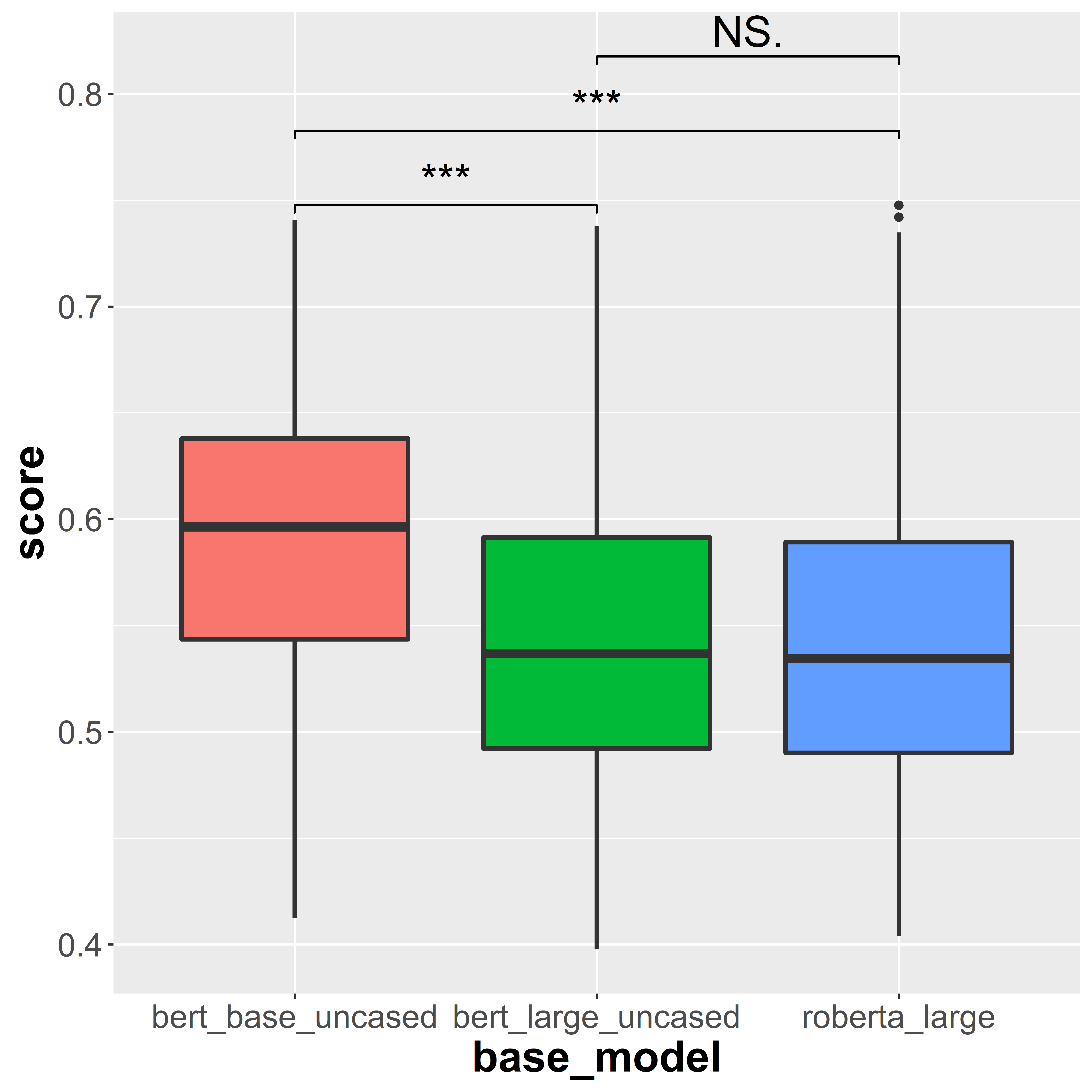}
         \caption{Plausibility:TSE}
         \label{tse_size}
     \end{subfigure}
     \hfill
     \begin{subfigure}[b]{0.48\textwidth}
         \centering
         \includegraphics[width=\textwidth]{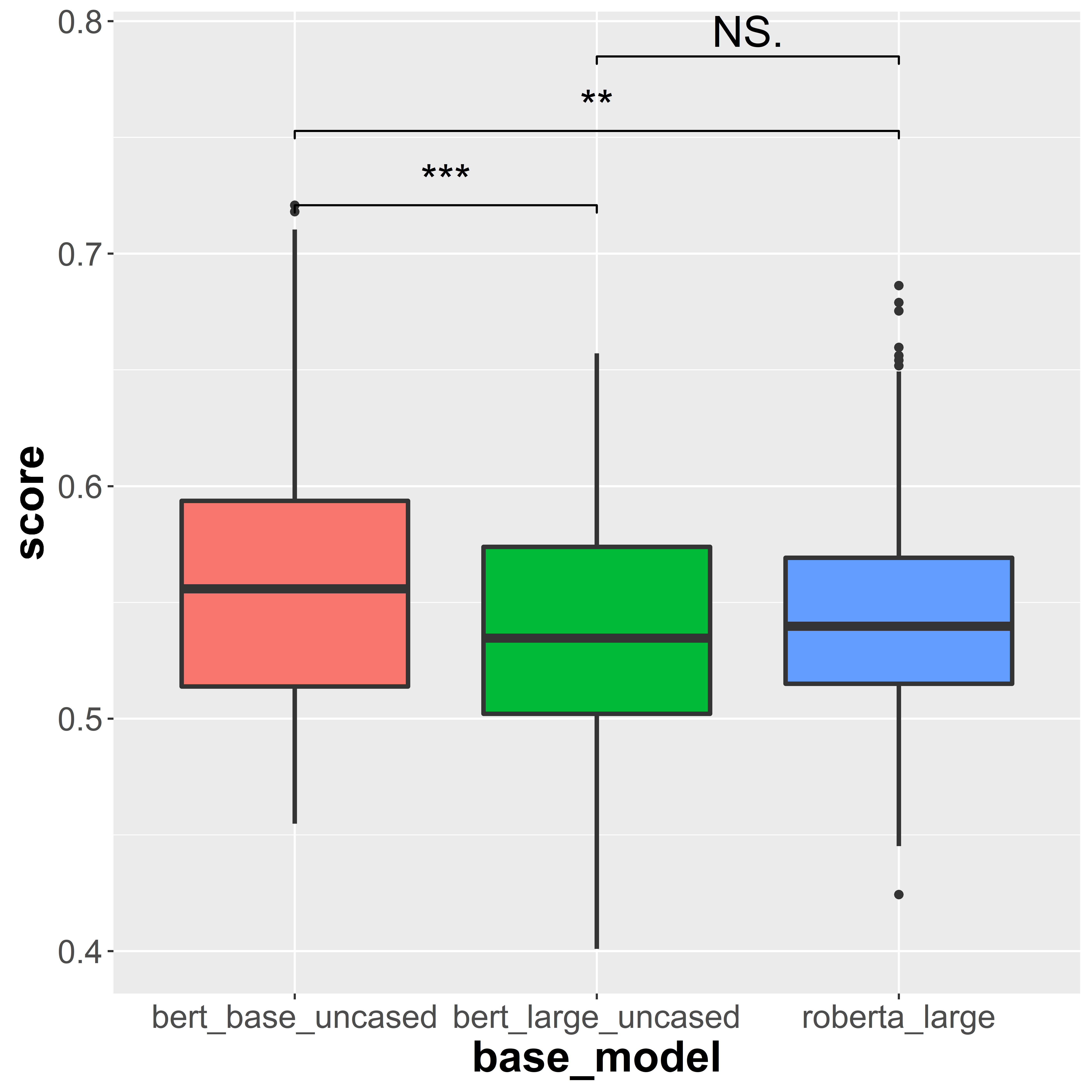}
         \caption{Plausibility:e-SNLI}
         \label{}
     \end{subfigure}

   \begin{subfigure}[b]{0.48\textwidth}
         \centering
         \includegraphics[width=\textwidth]{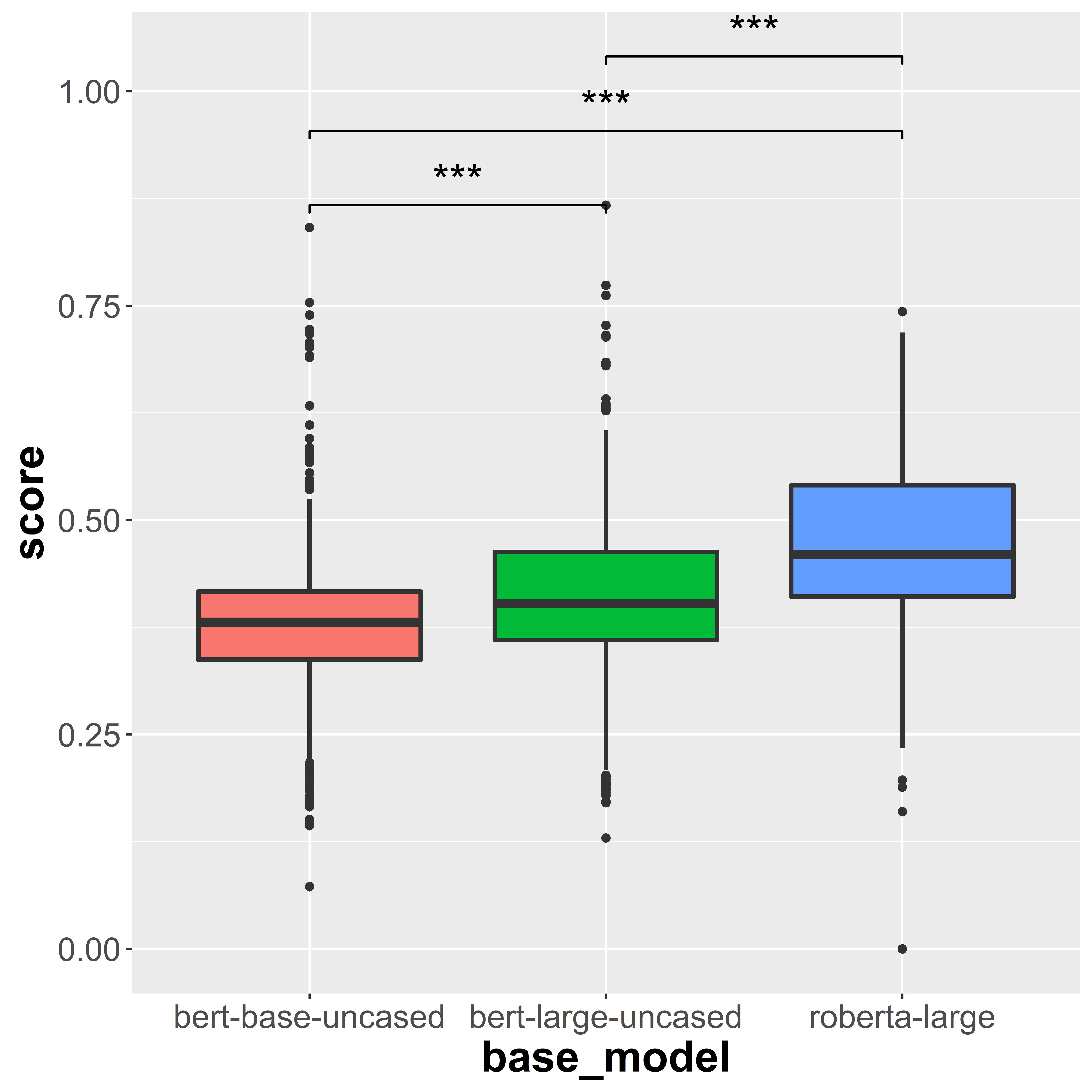}
         \caption{Faithfulness:TSE}
         \label{nli_size}
     \end{subfigure}
     \hfill
     \begin{subfigure}[b]{0.48\textwidth}
         \centering
         \includegraphics[width=\textwidth]{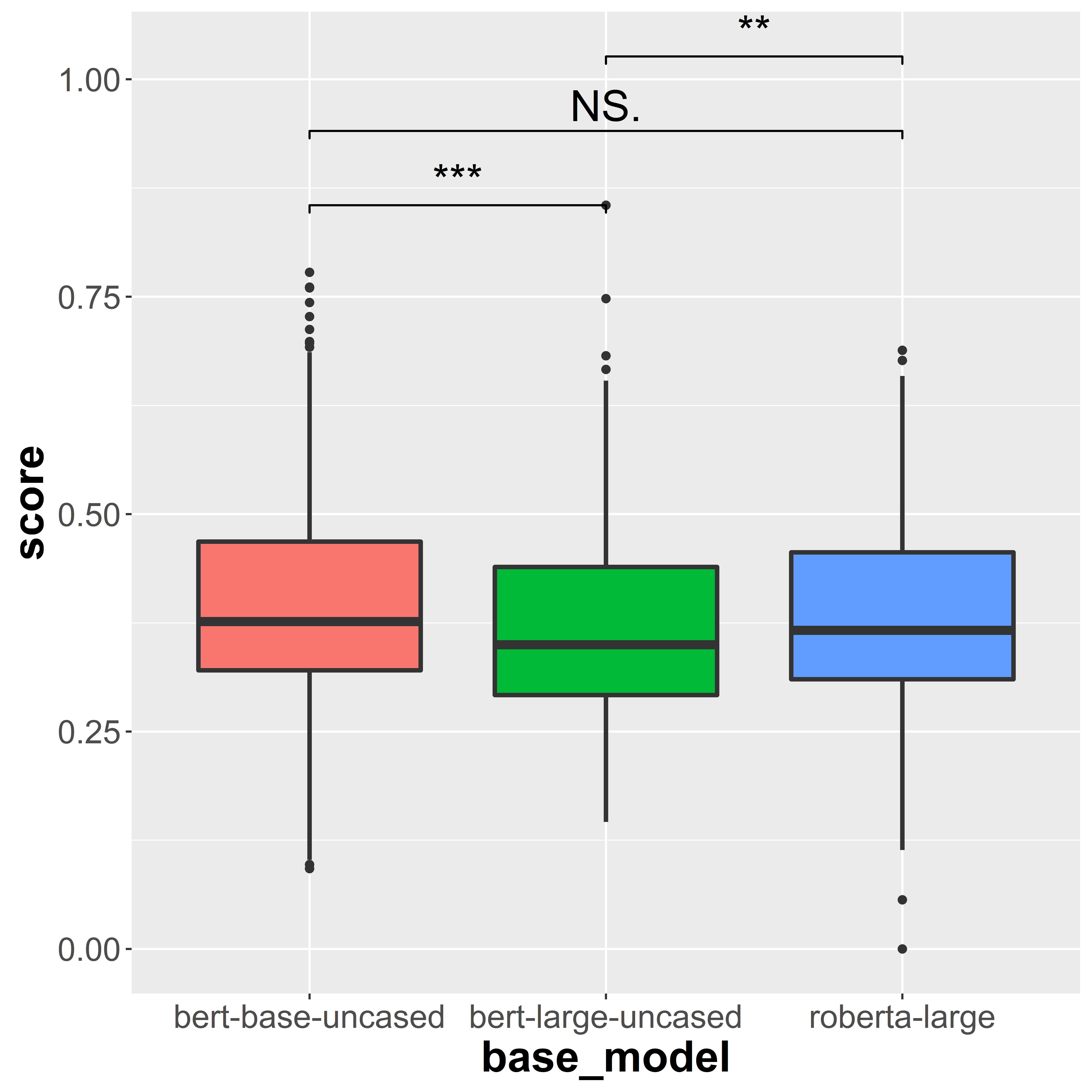}
         \caption{Faithfulness:e-SNLI}
         \label{nli_prompt}
     \end{subfigure}
     
  \caption{The Plausibility scores of base models, averaged across saliency methods, training sizes and prompting methods. \textit{attn} stands for attention. \textit{ig} stands for Integrated Gradients and \textit{shap} stands for Shapley Value Sampling. \textit{gold} stands for the gold annotations.\textit{NS} stands for the no significant difference. \textit{*, **, ***} stand for \textit{p}-value <.05, .01 and .001.}
    \label{compare_base_model}
\end{figure*}

Figure \ref{compare_sal} illustrates the plausibility and faithfulness scores per saliency method, averaged across models, training sizes, prompting methods and seeds. 
Figure \ref{compare_base_model} illustrates the plausibility scores per base model, averaged across saliency methods, training sizes, prompting methods and seeds.

\end{document}